\documentclass{article}
\usepackage{ijcai26}

\usepackage{multirow}
\usepackage{times}
\usepackage{soul}
\usepackage{url}
\usepackage[hidelinks]{hyperref}
\usepackage[utf8]{inputenc}
\usepackage[small]{caption}
\usepackage{graphicx}
\usepackage{amsmath}
\usepackage{amsthm}
\usepackage{booktabs}
\usepackage{algorithm}
\usepackage{algorithmic}
\usepackage[switch]{lineno}
\usepackage{amssymb}

\title{LISA: Language-guided Interference-aware Spatial-Frequency Attention for Driver Gaze Estimation}

\author{
Jun Ma$^{1,2}$
\and
Zhenye Yang$^{1,2}$
\and
Ruichen Zhou$^{1,2}$
\and
Pei Zhang$^{3}$
\and
Huan Li$^{4}$
\And
Jinpeng Chen$^{1,2}$\thanks{Corresponding author.}
\\
\affiliations
$^{1}$School of Computer Science (National Pilot Software Engineering School), Beijing University of Posts and Telecommunications\\
$^{2}$Key Laboratory of Trustworthy Distributed Computing and Service (BUPT), Ministry of Education\\
$^{3}$School of Electrical Engineering, Guangxi University\\
$^{4}$Zhejiang University\\
\emails
\{majun, yangzhenye, 2025010344\}@bupt.edu.cn,
pzhang@gxu.edu.cn,
lihuan.cs@zju.edu.cn,
jpchen@bupt.edu.cn
}

\begin{document}

\maketitle

\begin{abstract}
    Driver gaze estimation serves as a fundamental metric for evaluating driver attentiveness in modern monitoring systems. Beyond being vulnerable to sudden lighting changes and sensor noise, spatial-domain models struggle to disentangle authentic gaze cues from irrelevant visual attributes. In this paper, we propose LISA, a \textbf{L}anguage-guided \textbf{I}nterference-aware \textbf{S}patial-Frequency \textbf{A}ttention framework that combines frequency-domain priors with vision-language knowledge. Observing that the amplitude spectrum remains relatively stable even under spatial perturbations, we design a dual-domain fusion mechanism. It integrates stable low-frequency semantics into high-frequency details, employing spatial attention to precisely target ocular regions. To reduce semantic ambiguity, we also introduce a training-time disentanglement strategy. Using a frozen CLIP encoder and orthogonal regularization, we explicitly separate gaze features from appearance interference. Experiments on two benchmarks show that LISA achieves state-of-the-art performance, with significantly improved robustness against occlusions and lighting variations. The code repository is available at https://github.com/Mason-bupt/LISA. 
\end{abstract}

\section{Introduction}

Driver distraction and fatigue are among the major contributors to global traffic accidents, motivating the deployment of effective driver monitoring systems (DMS) to improve road safety \cite{rangesh2020driver,guo2024tla}. Among various behavioral cues, gaze direction is a key indicator of driver cognitive state and attention concentration \cite{hu2024context}. Although early solutions relied on supportive devices such as eye tracking glasses or specialized infrared sensors, the field has clearly shifted towards using standard RGB cameras for appearance-based gaze estimation \cite{liu2021generalizing,cheng2022puregaze}. These data-driven approaches, powered by Convolutional Neural Networks (CNNs) and Transformers, offer a non-invasive and cost-effective solution suitable for deployment in modern smart cockpits.

However, the transition of these models from a controlled environment to a real driving scenario poses many challenges. The visual environment in the vehicle is inherently unstable and drivers frequently encounter extreme lighting conditions, from dazzling sunlight to dimly lit tunnels \cite{li2024domain}. This complexity is further compounded by various physical attributes and occluding accessories, such as sunglasses and masks \cite{muhammad2020deep}.

In the presence of such visual instability, mainstream methods primarily employ deep convolutional backbone networks or visual transformers to extract high-level semantic features. These methods rely on the aggregate of local pixel patterns or patch-based markers to regress gaze vectors \cite{kellnhofer2019gaze360,cheng2022gaze,cheng2024you}. However, a critical limitation persists: these architectures operate primarily in the spatial domain. Consequently, they are inherently sensitive to pixel-level interference. When visual conditions change, the spatial distribution of key ocular features shifts unpredictably, causing the models to produce jittery or erroneous predictions \cite{zhang2015appearance,cheng2024appearance}. Moreover, without explicit disentanglement mechanisms, they tend to overfit irrelevant appearance attributes (e.g., associating ``sunglasses'' with a fixed gaze bias) rather than learning the true gaze feature \cite{cheng2022puregaze,kim2024appearance}.
\begin{figure}[htbp]
 \centering
 \includegraphics[width=\linewidth]{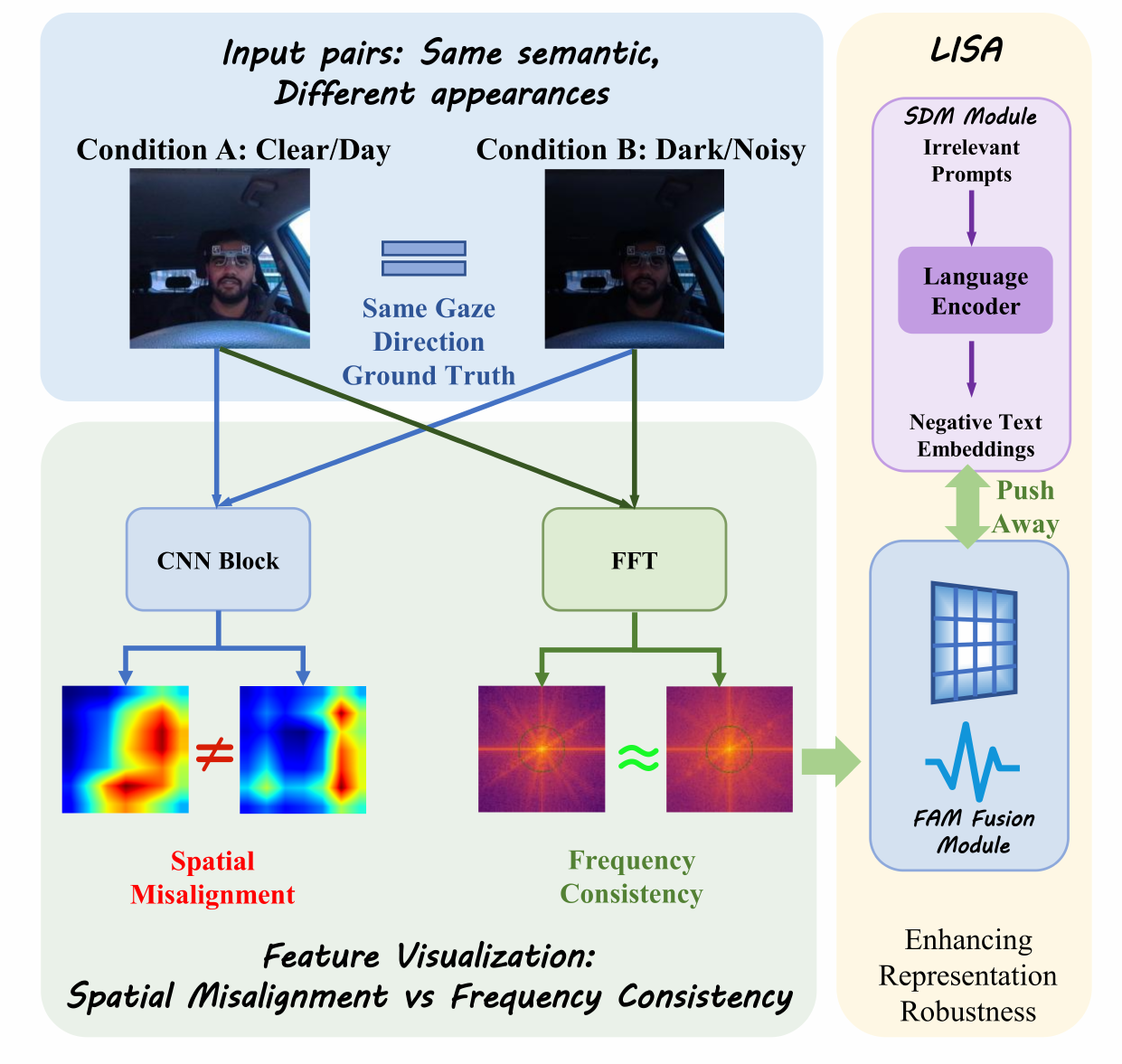}
 \caption{Visualizing the robustness gap. \textbf{Left:} Spatial features fail to align under lighting changes, while frequency spectra remain stable. \textbf{Right:} LISA enhances robustness by injecting frequency priors and employing a language-guided ``Push Away'' mechanism to suppress irrelevant semantic attributes.}

 \label{fig:motivation}
\end{figure}

This vulnerability to environmental shifts exposes the basic limitation of the current CNN-based paradigm, which we define as \textbf{Spatial Misalignment}. As illustrated in Figure~\ref{fig:motivation}, spatial features are highly sensitive to disturbances at the pixel-level. Under the condition of weak light or noise (Condition B), the feature distribution of the human eye region is significantly more degraded than that under the clear condition (Condition A), which makes it difficult for the model to capture subtle fixation clues.

Although the spatial pixel values fluctuate greatly, we observe that the overall structure of the face remains basically unchanged in the frequency domain. As shown in the frequency path of Figure~\ref{fig:motivation}, the amplitude spectrum shows \textbf{Frequency Consistency}, which can maintain a stable structural characteristic \cite{chen2021amplitude} even when the spatial image is corrupted. This observation prompted us to explore the frequency-domain prior as a stabilizer for gaze estimation.

To address the dual challenges of environmental instability and appearance entanglement, we propose LISA (\textbf{L}anguage-guided \textbf{I}nterference-aware \textbf{S}patial-Frequency \textbf{A}ttention), a robust framework for unconstrained driver gaze estimation. Our method is based on two core components. First, to mitigate environmental shifts, we introduce the FAM Fusion (Frequency-Attention Modulated Fusion) module within a ResNet-18 backbone \cite{he2016deep}. Unlike traditional spatial fusion, FAM Fusion incorporates a Spectral Injection Block that converts features into the frequency domain through the Fast Fourier Transform (FFT). This mechanism can isolate stable low-frequency semantic components and adaptively inject them into high-frequency details, thereby maintaining frequency consistency when spatial features are affected by lighting or noise. As a supplement to this spectrum alignment, we integrate a Spatial Saliency Gating mechanism. This spatial attention module explicitly reweights the feature map to highlight key eye regions, ensuring the precise localization of visual cues. Second, to resolve appearance entanglement, we propose a Semantic Disentanglement Module (SDM). We implemented a feature separation training strategy using cross-modal knowledge from visual language models (specifically a frozen CLIP encoder \cite{radford2021learning}). As depicted by the \textbf{Push Away} mechanism in Figure~\ref{fig:motivation}, this forces the model to minimize the similarity between gaze features and text embeddings of common distractors (e.g., ``a driver wearing sunglasses''), resulting in pure gaze representations that are orthogonal to appearance noise.

Our contributions are as follows:
\begin{itemize}

\item We propose LISA, a Language-guided Interference-aware Spatial-Frequency Attention framework, which is anchored in the concept of frequency consistency to tackle persistent spatial misalignment bottleneck caused by environmental shifts.

\item Our method integrates the FAM Fusion module to obtain stable spectral semantics and the SDM module to filter appearance-based ambiguity, achieving excellent performance in multiple benchmark tests.

\end{itemize}

\section{Related Work}

\subsection{Gaze Estimation}

Monitoring driver gaze direction can effectively reduce the incidence of road accidents \cite{li2015gaze,zhang2017mpiigaze,wan2021pupil,wan2021robust}. Since gaze direction is difficult to measure directly, previous work has transformed gaze estimation into a sophisticated 3D representation \cite{xiao20252gaze,kawana2025ga3ce}. Furthermore, certain methodologies incorporate additional features, such as synthetic 3D annotations \cite{bao2025gazegene}, 3D priors with weak supervision \cite{Cheng_2025_CVPR,Vuillecard_2025_CVPR}, and large-scale encoders for target estimation \cite{Ryan_2025_CVPR}. Generally, the incorporation of additional information can significantly improve the model's predictive accuracy.

However, gaze is inherently a continuous spatial behavior, and capturing the temporal dynamics can enhance the accuracy of  the estimation. Certain methodologies employ LSTMs \cite{kellnhofer2019gaze360} and attention mechanisms \cite{jindal2024spatio} to implicitly model temporal relevance. \cite{Yu_2025_CVPR} and \cite{Mazzamuto_2025_CVPR} proposed approaches that capture gaze dynamics, thereby enhancing video gaze estimation capabilities. The construction of a gaze estimation model encounters challenges such as disentangling authentic gaze cues from visual clutter. Additionally, \cite{du2022freegaze} and \cite{yin2024lg} proposed frameworks based on frequency domains and prompts.

\subsection{Gaze Estimation for Driver}

While invasive active sensors have been extensively researched, there has been a recent shift towards non-intrusive vision-based systems. Existing studies often approximated gaze by estimating sparse zones \cite{yang2021driver,hu2024context}. Consequently, with the advent of modern benchmarks, researchers begin to explore how to predict accurate 3D direction continuously. \cite{kasahara2022look} and \cite{cheng2024you} proposed frameworks based on self-supervision and dual-stream Transformers to capture complex line-of-sight patterns. In particular, FIFA \cite{Hu_2025_CVPR} introduces a fine-grained inter-frame attention module designed to track subtle pupil movements over time. However, the susceptibility of spatial domain methods to environmental changes restricts their applicability. 

\section{Methodology}
In this section, we provide a comprehensive delineation of the proposed LISA framework. In detail, first, we introduce the problem formulation. Second, we provide an overview of the architecture. Next, we detail the core FAM Fusion module and the SDM. Finally, we describe the Regression Head and optimization objectives.

\begin{figure*}[t]
 \centering
 \includegraphics[width=\textwidth]{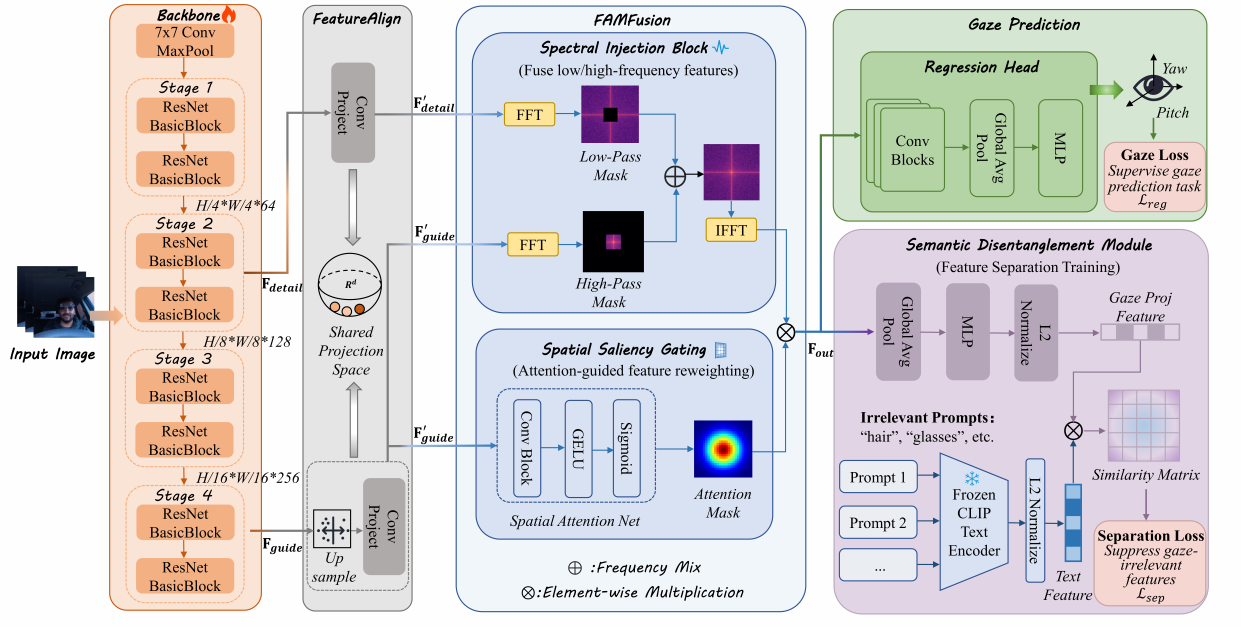}
 \caption{Overview of the LISA framework. The architecture consists of a ResNet-18 backbone extracting multi-scale features, a Feature Alignment module for dimension unification, the core FAM Fusion module, and the Semantic Disentanglement Module.}
 \label{fig:overall_architecture}
\end{figure*}

\subsection{Problem Formulation}
Specifically, we utilize a single RGB input image $\mathbf{X} \in \mathbb{R}^{3 \times H \times W}$ depicting a driver to estimate the gaze direction. Considering that ground truth annotations in standard gaze datasets are typically provided as 3D unit vectors $\mathbf{v} = (x, y, z) \in \mathbb{R}^3$, we formulate the regression task using 2D spherical coordinates, specifically predicting the angles of yaw ($\phi$) and pitch ($\theta$), i.e., $\mathbf{g} = (\phi, \theta) \in \mathbb{R}^2$. Since the 2D pitch and yaw angles are sufficient to uniquely describe the gaze direction within the head coordinate system, we adopt this 2D representation. Additionally, in order to be generally more stable and easier for the model to learn compared to a 3D vector constrained to the unit sphere, we regress a 2D continuous variable. This conversion between the 3D unit vector $\mathbf{v}$ and the 2D angles $\mathbf{g}$ follows standard spherical coordinate transformations. As a result, it allows for seamless integration with 3D evaluation metrics.

\subsection{Framework Overview}
As illustrated in Figure~\ref{fig:overall_architecture}, the proposed LISA framework is built on a lightweight ResNet backbone. Specifically, the backbone extracts multi-scale feature maps: Detail Features $\mathbf{F}_{detail}$ from the shallow layer, which preserves high-frequency spatial cues. Meanwhile, the deep features $\mathbf{F}_{guide}$ of the deep layer contain a rich semantic context. The aligned features $\mathbf{F}'_{detail}$ and $\mathbf{F}'_{guide}$ are fed into the FAM Fusion module, where they are fused via spectral and spatial mechanisms. The fused feature is then processed by a Regression Head to predict the gaze vector $\mathbf{g}$. To facilitate the learned features to be robust to appearance interference, we propose an auxiliary Semantic Disentanglement Module, which leverages a frozen CLIP encoder to regularize the latent space.

\subsection{Feature Alignment}
In order to align different scales and channel dimensions of the multi-level features extracted by the backbone prior to the core fusion process, we employ a Feature Alignment block to unify these properties. Consequently, the detail and guide features can be processed effectively in subsequent modules. We denote the aligned features by $\mathbf{F}'_{detail}$ and $\mathbf{F}'_{guide}$.

The alignment process is formulated as follows.
\begin{equation}
\mathbf{F}'_{detail} = \phi_{align}(\mathbf{F}_{detail})
\end{equation}
\begin{equation}
 \mathbf{F}'_{guide} = \text{Upsample}(\phi_{align}(\mathbf{F}_{guide}))
 \end{equation}
Specifically, $\phi_{align}$ includes a $1 \times 1$ convolution followed by Batch Normalization and GELU activation; the projection maps both feature streams to a unified channel dimension $C$. Additionally, in order to match the spatial resolution of the detail features $\mathbf{F}_{detail}$, we apply bilinear upsampling to the deep features $\mathbf{F}_{guide}$.

\subsection{Frequency-Attention Modulated Fusion (FAM Fusion)}
Traditional methods still suffer from ``Spatial Misalignment'', owing to their heavy reliance on unstable spatial pixels. As the environment changes, the issue worsens markedly. To bridge this gap, we propose a Frequency-Attention Modulated Fusion, called FAM Fusion. To facilitate the network in guaranteeing frequency consistency, we design a Spectral Injection Block. Furthermore, a Spatial Saliency Gating mechanism is integrated to sharpen local attention, thereby specifically improving focus.

\subsubsection{Spectral Injection Block}
We introduce a Spectral Injection Block aimed at extracting semantic priors from the deep branch and replacing the low-frequency components of the detail branch.

\paragraph{FFT Transformation.}
We map spatial features to the frequency domain using a 2D real-to-complex Fast Fourier Transform (rFFT), denoted as $\mathcal{F}_{r}$. Specifically, $\mathcal{F}_{r}$ encodes the aligned detail and guide features $\mathbf{F}'_{detail}$ and $\mathbf{F}'_{guide}$ into complex-valued spectral matrices $\mathbf{Y}_{detail}$ and $\mathbf{Y}_{guide}$:
\begin{equation}
    \mathbf{Y}_{detail} = \mathcal{F}_{r}(\mathbf{F}'_{detail})
\end{equation}
\begin{equation}
    \mathbf{Y}_{guide} = \mathcal{F}_{r}(\mathbf{F}'_{guide})
\end{equation}
where $\mathcal{F}_{r}$ denotes the rFFT operation, and $\mathbf{Y} \in \mathbb{C}^{C \times H \times (\lfloor W/2 \rfloor + 1)}$ represents the complex-valued spectral matrix.

\paragraph{Adaptive Spectral Mixing.}We define a binary mask matrix $\mathbf{M}$ to isolate the low-frequency region determined by a ratio $\gamma$. A learnable scalar parameter $\alpha \in (0, 1)$ is used as an adaptive weight for spectral mixing. The fused spectrum $\mathbf{Y}_{fused}$ is obtained as follows:
\begin{equation}
    \mathbf{Y}_{fused} = \mathbf{Y}_{detail} \cdot (1 - \alpha \cdot \mathbf{M}) + \mathbf{Y}_{guide} \cdot (\alpha \cdot \mathbf{M})
\end{equation}
where $\mathbf{M}$ is a binary mask matrix isolating the low-frequency region determined by a ratio $\gamma$, and $\alpha \in (0, 1)$ is a scalar parameter that can be learned by constraining a sigmoid function. This framework extracts semantic priors from the deep stream and assigns weights to them, thereby preserving high-frequency edge information from $\mathbf{Y}_{detail}$ and substituting its low-frequency components with $\mathbf{Y}_{guide}$. 

Finally, the inverse rFFT operation recovers the spatial feature $\mathbf{F}_{freq}$:
\begin{equation}
    \mathbf{F}_{freq} = \mathcal{F}^{-1}_{r}(\mathbf{Y}_{fused})
\end{equation}
where $\mathcal{F}^{-1}_{r}$ denotes the inverse rFFT operation.
\subsubsection{Spatial Saliency Gating}
The representation of spatial features is influenced by the coupling factors of global structural stability and background noise. Therefore, we employ Spatial Saliency Gating to capture the ocular region information, thereby suppressing background noise.

Specifically, a lightweight network is employed to extract attention features from the deep features $\mathbf{F}'_{guide}$. We generate a spatial attention matrix $\mathbf{A} \in \mathbb{R}^{1 \times H \times W}$ as follows:
\begin{equation}
\mathbf{A} = \sigma(\text{Conv}_2(\delta(\text{Conv}_1(\mathbf{F}'_{guide}))))
\end{equation}
where $\mathbf{A} \in \mathbb{R}^{1 \times H \times W}$ is the attention map, $\sigma$ is the Sigmoid function, $\delta$ is the GELU activation, and $\text{Conv}_{1,2}$ are convolutional layers. 

Finally, the output features are computed as follows:
\begin{equation}
\mathbf{F}_{out} = \mathbf{F}_{freq} \cdot (\mathbf{A} + \epsilon)
\end{equation}

Here, $\epsilon$ indicates a residual scalar bias term used to stabilize training in its early stages.

\subsection{Semantic Disentanglement Module (SDM)}
In order to resolve the Semantic Ambiguity where gaze features are entangled with appearance attributes, we need to ensure that the learned representations contain orthogonal regularization. To achieve this goal, we utilize the rich semantic knowledge of a pre-trained Vision-Language Model and introduce a Semantic Disentanglement Module (SDM).

\paragraph{Text-Driven Negative Anchors.}Specifically, for a given set of ``negative'' semantic anchors that represent gaze-irrelevant appearance factors, where $T$ is the prompt pool, the diverse appearance attributes are expressed as a set of $N$ prompts $T = \{t_1, t_2, \dots, t_N\}$ through natural language templates. In detail, we freeze the CLIP Text Encoder and encode these prompts to obtain a semantic embedding matrix $\mathbf{E}_{text} \in \mathbb{R}^{N \times D}$.

\paragraph{Feature Projection.}To elaborate, we project the fused spatial feature $\mathbf{f}_{gaze}$ to an intermediate representation $\mathbf{e}_{gaze}$ with a projection head $\phi_{proj}$. A Linear layer is then employed to project these feature vectors into a shared CLIP latent space:
\begin{equation}
 \mathbf{e}_{gaze} = \phi_{proj}(\mathbf{f}_{gaze}) \in \mathbb{R}^D
\end{equation}
where $\mathbf{e}_{gaze}$ is the projected visual embedding.

\paragraph{Orthogonal Regularization.}Since the main objective is to ensure that the gaze representation $\mathbf{e}_{gaze}$ contains no information related to the appearance attributes defined in $\mathbf{E}_{text}$, we hope that the optimizer can minimize any correlation between the gaze feature and the appearance text embeddings. By calculating the cosine similarity between the projected gaze feature and each text embedding and applying the mean operation, we obtain the separation loss $\mathcal{L}_{sep}$, which encodes the correlation between the gaze feature and the appearance text embeddings. The separation loss is defined as:
\begin{equation}
    \mathcal{L}_{sep} = \frac{1}{N} \sum_{i=1}^{N} \left| \frac{\mathbf{e}_{gaze} \cdot \mathbf{E}_{text, i}}{\|\mathbf{e}_{gaze}\| \|\mathbf{E}_{text, i}\|} \right|
\end{equation}
where $|\cdot|$ denotes the absolute value operation, ensuring that the optimizer minimizes any correlation (positive or negative) between the gaze feature and the appearance text embeddings $\mathbf{E}_{text, i}$.

\subsection{Regression Head and Loss Function}

\paragraph{Coordinate-aware Regression.}Specifically, for a given pair of generated normalized coordinate maps $\mathbf{C}_x, \mathbf{C}_y \in \mathbb{R}^{H \times W}$, where $(h, w)$ is the spatial location, the position information is represented as a value defined as:
\begin{equation}
    \mathbf{C}_x(h, w) = \frac{2w}{W-1} - 1
\end{equation}

\begin{equation} 
\mathbf{C}_y(h, w) = \frac{2h}{H-1} - 1
\end{equation}

Normalized spatial coordinates are concatenated according to channel dimension, resulting in a complete feature map $\mathbf{F}_{coord}$:
\begin{equation}
    \mathbf{F}_{coord} = \text{Concat}(\mathbf{F}_{out}, \mathbf{C}_x, \mathbf{C}_y)
\end{equation}

The regression head consists of a Multi-Layer Perceptron (MLP) aimed at predicting final gaze angles from the augmented feature $\mathbf{F}_{coord}$ and pooling it across spatial locations:
\begin{equation}
    \hat{\mathbf{g}} = \Phi_{reg}(\text{Pool}(\mathbf{F}_{coord}))
\end{equation}
where $\Phi_{reg}$ denotes the MLP regression network, and $\hat{\mathbf{g}} = (\hat{\phi}, \hat{\theta})$ represents the predicted yaw and pitch angles.

\paragraph{Total Objective.} We use the following composite loss function to minimize our total objective:
\begin{equation}
\mathcal{L}_{total} = \mathcal{L}_{reg} + \lambda_{sep} \mathcal{L}_{sep}
\end{equation}

To strictly constrain the geometric consistency of the gaze direction, a regression loss $\mathcal{L}_{reg}$ is used to combine the Smooth L1 loss and the Angular Loss, thus constraining the geometric consistency of the predicted 3D vector:
\begin{equation}
\mathcal{L}_{reg} = \mathcal{L}_{L1}(\mathbf{g}, \hat{\mathbf{g}}) + \lambda_{ang} \mathcal{L}_{ang}(\mathbf{v}, \hat{\mathbf{v}})
\end{equation}
where $\mathcal{L}_{L1}$ denotes the loss in pitch and yaw angles, and $\mathcal{L}_{ang}$ represents the angular error in degrees between the predicted 3D vector $\hat{\mathbf{v}}$ and the ground truth vector $\mathbf{v}$.

\section{Experiments}

\subsection{Datasets}
We used the following datasets in our experiments: LBW \cite{kasahara2022look} and IVGaze \cite{cheng2024you}.

LBW is a large-scale dataset that includes driver facial images, road-facing scene images, and 3D gaze directions from head-mounted eye trackers. Facial landmarks are used to crop facial images with a fixed resolution size. The dataset contains 28 subjects. We divided the dataset into two subsets based on subjects, with subjects with IDs 1–22 as a training set and the rest as a test set.

The IV dataset comprises 44,703 images in 125 subjects. It encompasses a rich set of ground truth data, including head pose and gaze direction. We adhere to the dataset division strategy outlined in \cite{cheng2024you} and implement a three-fold cross-validation approach.

\subsection{Implementation Details}
The proposed method is implemented using PyTorch and trained for 100 epochs on an NVIDIA RTX 4090 GPU. A batch size of 64 is used for training, and the learning rate is set to $1 \times 10^{-4}$. Optimization is performed using the AdamW optimizer with a weight decay of $5 \times 10^{-4}$.

\subsection{Comparison with SOTA Methods}
We compare our approach with the following baselines: (i) ResNet-18 \cite{he2016deep}, (ii) Gaze360 \cite{kellnhofer2019gaze360}, (iii) XGaze \cite{zhang2020eth}, (iv) GazeTR \cite{cheng2022gaze}, (v) GazePTR \cite{cheng2024you}, (vi) STAGE \cite{jindal2024spatio} and (vii) FIFA \cite{Hu_2025_CVPR}. All the compared baselines rely on CNN- or Transformer-based (or hybrid) feature extractors to learn appearance representations for gaze estimation.

The mean angle error is used to evaluate the accuracy of the model prediction, with lower values representing better performing methods. 
\begin{table}
    \centering
    \begin{tabular}{lcccc}
        \toprule
        \multirow{2}{*}{Method} & \multicolumn{2}{c}{Error} & \multicolumn{2}{c}{Specification} \\
        
        \cmidrule(lr){2-3} \cmidrule(lr){4-5}
        
         & IV & LBW & FLOPs & \#Param. \\
        \midrule
        
        Gaze360  & $8.23^\circ$ & $7.12^\circ$ & 12.78G & 14.60M \\
        ResNet-18  & $8.02^\circ$ & $6.41^\circ$ & 3.65G  & 11.25M \\
        GazeTR    & $7.33^\circ$ & $6.17^\circ$ & 3.68G  & 11.39M \\
        XGaze     & $7.35^\circ$ & $6.23^\circ$ & 8.26G  & 23.64M \\
        GazePTR   & $7.09^\circ$ & $6.05^\circ$ & 3.69G  & 11.52M \\
        STAGE     & $7.34^\circ$ & $6.12^\circ$ & 3.67G  & 11.32M \\
        FIFA     & $7.29^\circ$ & $5.84^\circ$ & $\mathbf{2.70G}$  & 5.87M \\
        
        Ours          & $\mathbf{7.05}^\circ$ & $\mathbf{5.39}^\circ$ & 5.65G & $\mathbf{4.39M}$ \\
        \bottomrule
    \end{tabular}
    \caption{Performance comparison. Best results are marked in \textbf{bold}.}
    \label{tab:comparison}
\end{table}

In Table~\ref{tab:comparison}, we present a comparative analysis of the performance of our method against state-of-the-art techniques. Initially, in the IV dataset, our method achieves an improvement with an error of $7.05^\circ$ representing a reduction of $0.97^\circ$ relative to the ResNet-18's error of $8.02^\circ$. The performance of our proposed method also surpasses the recent hybrid architecture FIFA ($7.29^\circ$) by a margin of $0.24^\circ$. For the LBW dataset, our method similarly exhibits superior performance, achieving an error of $5.39^\circ$, which signifies a decrease of $1.02^\circ$ from ResNet-18's error of $6.41^\circ$. Compared to FIFA ($5.84^\circ$), our proposed method enhances performance by 7.7\%, demonstrating that our method outperforms current methods in terms of accuracy to ensure high precision and robustness in complex driving scenarios challenged by various visual interferences.

For FLOPs, our method requires 5.65G operations to support frequency-domain modeling, representing a moderate computational cost for improved accuracy. Moreover, our method encompasses only 4.39M parameters, a significantly smaller number compared to Gaze360's 14.6M and XGaze's 23.64M, rendering the model more suitable for deployment in resource-limited settings. 

\subsection{Visualization of Prediction Results}
\begin{figure}[t]
 \centering
 \includegraphics[width=\linewidth]{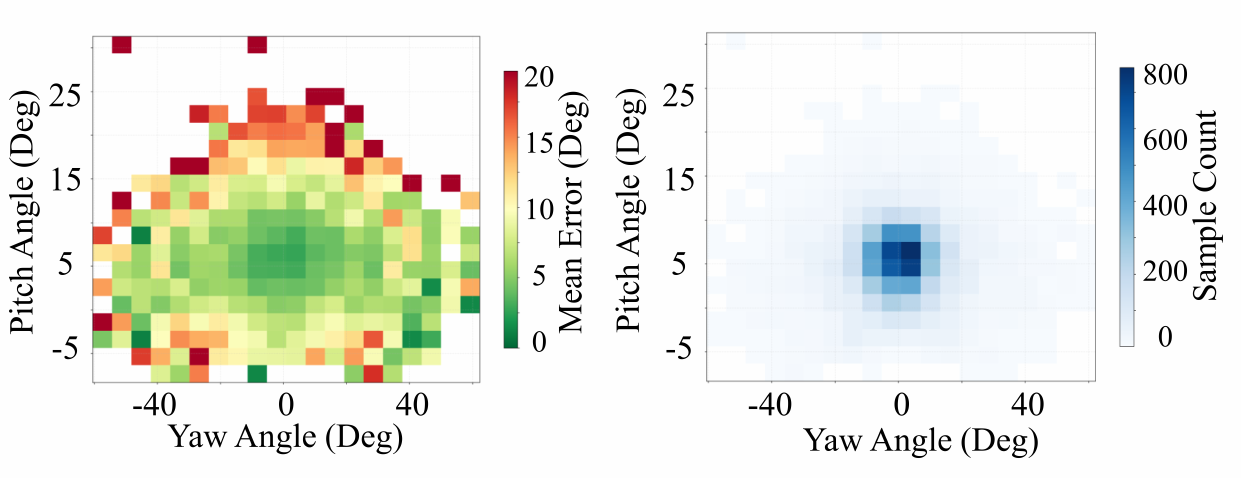}
 \caption{Error and sample distribution in gaze angle space. The x-axis denotes yaw angle, and the y-axis denotes pitch angle.}
 \label{fig:dis}
\end{figure}

When driving a vehicle, the gaze direction is usually fixed in a few areas. Figure~\ref{fig:dis} highlights the utilization of the average precision within the pitch and yaw dimensions to assess the model's performance at varying levels of permitted error. The MAE heatmap (Left) shows an expansive high-precision region, maintaining consistently low error rates where the pitch range is $[-5^\circ, 15^\circ]$ and the yaw range is $[-40^\circ, 40^\circ]$. This outcome ensures that a larger number of samples maintains prediction accuracy within the designated error limits, which is particularly beneficial for driver monitoring tasks. By cross-referencing with the sample density map (Right), we observe a matching accuracy in areas where samples are concentrated, and the proposed method can obtain a more accurate gaze estimation. Moreover, the model's performance exhibits a smooth error gradient rather than abrupt performance drops, suggesting a higher concentration of samples within these bounds. In general, these results demonstrate that LISA-based representations are effective and can be easily transferred to most real-world applications.

\subsection{Embedding Visualization}

\begin{figure}[t]
 \centering
 \includegraphics[width=\linewidth]{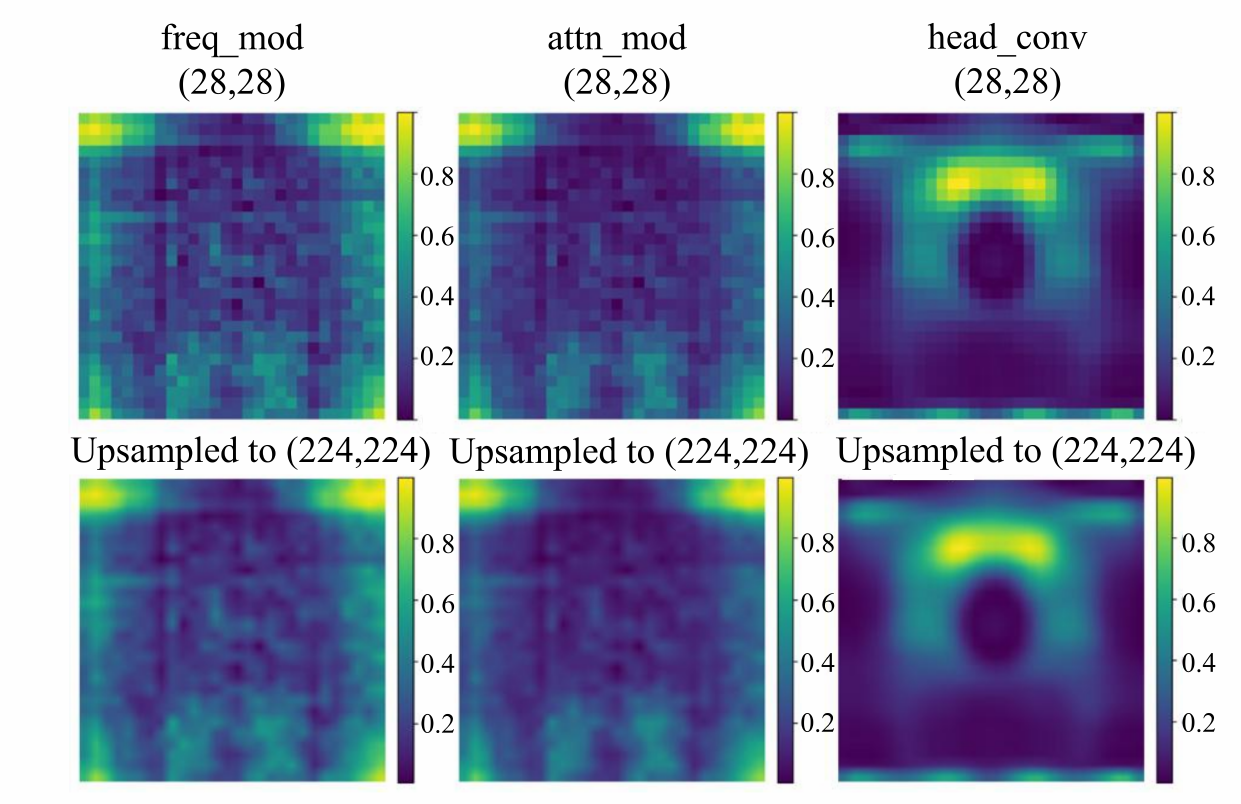}
 \caption{Visualization of multi-resolution feature evolution.}

 \label{fig:emb}
\end{figure}

To substantiate the efficacy of the proposed internal components, we present a visualization of the feature maps of the test dataset. Figure~\ref{fig:emb} illustrates the progressive evolution of feature maps in the LBW dataset through the \textit{Spectral Injection Block}, \textit{Spatial Saliency Gating} and the final Regression Head output. For the feature distributions of the initial activations in the evaluation datasets, it can be seen that the features exhibit distributed, texture-rich patterns, confirming that our frequency-aware fusion explicitly preserves fine-grained structural details. Crucially, we observe that the final output embeddings exhibit more refined distributions rather than a simple localized hotspot. This transformation demonstrates that the head successfully integrates spatial coordinates with global orientation cues to bridge the gap between abstract features and continuous gaze regression.

\begin{figure}[htbp]
    \centering
    \includegraphics[width=\linewidth]{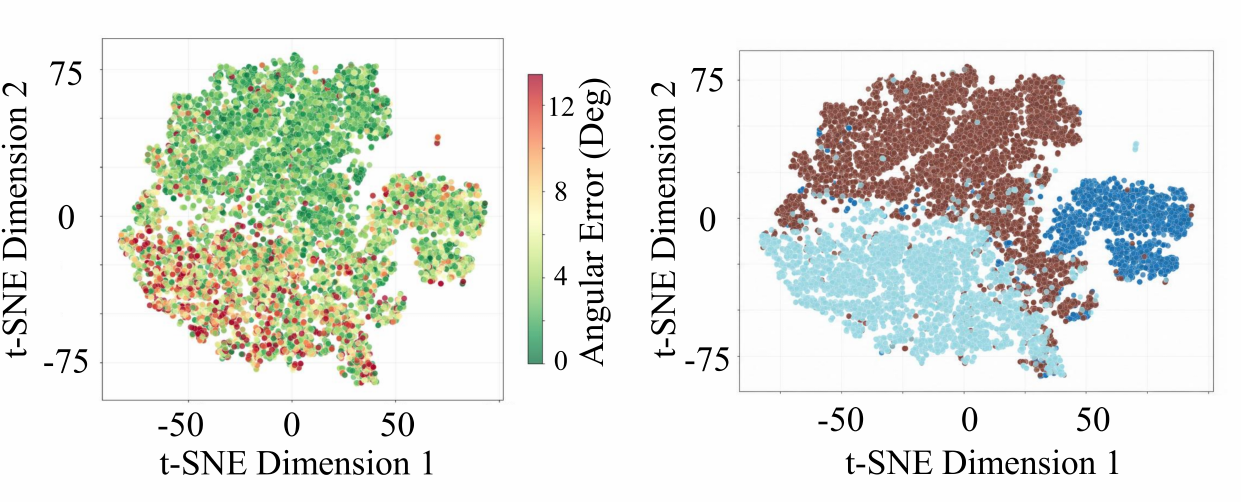} 
    \caption{t-SNE visualization; axes denote t-SNE dimensions, with colors indicating error and subject ID, respectively.}
    \label{fig:tsne}
\end{figure}
\begin{table*}[t]
    \centering
    \begin{tabular}{lc|cc|cc|cc}
        \toprule
        \multirow{2}{*}{Category} & \multirow{2}{*}{Count} & \multicolumn{2}{c}{Mean Error ($\downarrow$)} & \multicolumn{2}{c}{Std. Dev.} & \multicolumn{2}{c}{Accuracy ($< 8^\circ$ $\uparrow$)} \\
        \cmidrule(lr){3-4} \cmidrule(lr){5-6} \cmidrule(lr){7-8}
         & & GazePTR & \textbf{Ours} & GazePTR & \textbf{Ours} & GazePTR & \textbf{Ours} \\
        \midrule
        \textbf{Overall} & 44,703 & $7.09^\circ$ & $\mathbf{7.05}^\circ$  & $5.61^\circ$& $\mathbf{5.40}^\circ$ & 66.3\% & $\mathbf{66.8}\%$ \\
        \midrule
        \multicolumn{8}{l}{\textit{Impact of Accessories}} \\
        \hspace{1em} w/ Accessories & 30,720 & $7.30^\circ$ & $\mathbf{7.29}^\circ$ & $5.81^\circ$ & $\mathbf{5.49}^\circ$  & 64.8\% & $\mathbf{65.6}\%$ \\
        \hspace{1em} w/o Accessories & 13,983 & $6.66^\circ$ & $\mathbf{6.53}^\circ$ & $5.19^\circ$ & $\mathbf{5.11}^\circ$ & 69.1\% & $\mathbf{69.7}\%$ \\
        \midrule
        \multicolumn{8}{l}{\textit{Impact of Occlusions}} \\
        \hspace{1em} w/ Glasses & 27,774 & $7.22^\circ$ & $\mathbf{7.20}^\circ$
        & $5.62^\circ$  & $\mathbf{5.38}^\circ$  & 65.3\% & $\mathbf{65.8}\%$ \\
        \hspace{1em} w/o Glasses & 16,929 & $6.88^\circ$ & $\mathbf{6.76}^\circ$ 
        & $5.47^\circ$ & $\mathbf{5.34}^\circ$  & 67.3\% & $\mathbf{68.5}\%$ \\
        \hspace{1em} w/ Mask & 8,331 & $7.81^\circ$ & $\mathbf{7.54}^\circ$  & $6.31^\circ$ & $\mathbf{5.93}^\circ$& 63.2\% & $\mathbf{65.2}\%$ \\
        \hspace{1em} w/o Mask & 36,372 & $6.92^\circ$ & $\mathbf{6.92}^\circ$  & $5.41^\circ$ & $\mathbf{5.25}^\circ$ & 66.9\% & $\mathbf{67.2}\%$ \\
        \midrule
        \multicolumn{8}{l}{\textit{Impact of Environment}} \\
        \hspace{1em} Outdoor & 40,865 & $7.14^\circ$ & $\mathbf{7.12}^\circ$ & $5.66^\circ$  & $\mathbf{5.44}^\circ$ & 65.8\% & $\mathbf{66.3}\%$ \\
        \hspace{1em} Indoor & 3,838 & $\mathbf{6.00}^\circ$ & $6.06^\circ$  & $4.75^\circ$ & $\mathbf{4.48}^\circ$ & $\mathbf{73.4}\%$ & 73.1\% \\
        \bottomrule
    \end{tabular}
    \caption{Comparative robustness analysis between GazePTR model and our proposed LISA framework. Best results are marked in \textbf{bold}.}
    \label{tab:comparison_robustness}
\end{table*}

To rigorously validate the generalization capabilities of the model, we employ t-SNE to project high-dimensional feature embeddings into a 2D manifold, as visualized in Figure~\ref{fig:tsne}. 

We first examine the \textit{Prediction Error Distribution} (Left), which reveals a highly structured topology regarding the placement of correct and incorrect samples. Rather than a random scatter, the manifold exhibits a clear ``core-periphery'' organization: samples with high estimation precision (dense green regions) constitute the primary body of the distribution, while high-error samples (red/yellow) are largely marginalized to the sparse peripheries. This indicates that the model has successfully learned a robust mapping for the vast majority of the gaze distribution, effectively pushing failure cases—likely corresponding to extreme angles—to the geometric boundaries of the feature space.

Turning to the \textit{Subject ID Clustering} (Right), we observe that the embeddings naturally segregate into distinct clusters based on individual identities, confirming that the model preserves inherent appearance variations such as facial morphology. Crucially, by cross-referencing these two views, we find that the favorable error topology observed in the Left plot is consistently replicated within \textit{each} subject-specific cluster. Specifically, every identity cluster exhibits the same stable structure: a dense, accurate core surrounded by sparse, higher-error outliers. This structural consistency demonstrates that while the model encodes subject-specific differences (hence the separation), its gaze estimation mechanism remains fundamentally stable and unbiased across different individuals.
\begin{table}
    \centering
    \begin{tabular}{lcc}
        \toprule
        Method & IV Mean & LBW Mean \\
        \midrule
        ResNet-18 & $8.02^\circ$ & $6.41^\circ$ \\
        w/o Spectral Injection Block & $7.41^\circ$ & $6.33^\circ$ \\
        w/o Spatial Saliency Gating & $7.25^\circ$ & $5.88^\circ$ \\
        w/o SDM & $7.24^\circ$ & $5.54^\circ$ \\
        Ours & $\mathbf{7.05}^\circ$ & $\mathbf{5.39}^\circ$ \\
        \bottomrule
    \end{tabular}
    \caption{Ablation study results (Mean Error).}
    \label{tab:ablation}
\end{table}


\subsection{Robustness Analysis Against Occlusions and Environmental Factors}

We perform a 3-fold cross-validation on 44,703 IV dataset samples to evaluate LISA under unconstrained conditions. Table~\ref{tab:comparison_robustness} compares LISA with GazePTR on accessories, occlusions, and environmental shifts. LISA achieves a lower mean error of $\mathbf{7.05^\circ}$ and a reduced standard deviation ($\mathbf{5.40^\circ}$ vs. $5.61^\circ$). These results demonstrate superior stability and robustness across diverse driving scenarios.

\paragraph{Resilience to Accessories and Visual Clutter.}A broad analysis of the ``w/ Accessories'' category reveals the model's exceptional stability amidst visual distractions. Despite the introduction of diverse facial ornaments in 30,720 samples, our method maintains a competitive error of $7.29^\circ$. Performance degradation relative to the clean baseline (``w/o Accessories'') is limited to $0.76^\circ$. This minimal impact confirms that our feature extractor effectively disentangles gaze-relevant cues from extraneous appearance variations, preventing the model from overfitting to irrelevant facial ornaments.

\paragraph{Robustness Against Specific Occlusions.}
The model demonstrates remarkable resilience to common interferences. In scenarios involving eyewear (``w/ Glasses''), which often introduce specular reflections, our method achieves an error of $7.20^\circ$, outperforming GazePTR ($7.22^\circ$) with a marginal impact of only $0.44^\circ$ compared to subjects without glasses. This validates the efficacy of our Frequency Modulation module, which suppresses high-frequency glare artifacts. Notably, the advantage of our framework becomes more pronounced in severe occlusion scenarios such as facial masks (``w/ Mask''). Although GazePTR suffers a significant performance drop to $7.81^\circ$, our method maintains high precision at $\mathbf{7.54^\circ}$—a substantial improvement of $\mathbf{0.27^\circ}$. Furthermore, our method achieves a much lower standard deviation ($5.93^\circ$ vs. $6.31^\circ$), proving that the Attention Modulation mechanism successfully redirects focus to exposed ocular regions to compensate for the loss of lower-facial context.

\paragraph{Environmental Adaptability.}
The environmental analysis highlights the model's robustness to illumination changes. In the ``Outdoor'' setting, which accounts for the majority of wild samples, our model achieves an error of $7.12^\circ$, consistently outperforming GazePTR ($7.14^\circ$). While GazePTR shows a slight advantage in mean error under ideal ``Indoor'' conditions ($6.00^\circ$ vs. $6.06^\circ$), our method exhibits superior stability with a significantly lower standard deviation ($4.48^\circ$ vs. $4.75^\circ$). This indicates that the LISA framework delivers more consistent predictions in varying lighting conditions, ensuring reliability whether it is deployed in controlled indoor environments or harsh outdoor settings.

\subsection{Ablation Study}

To systematically investigate the contribution of each component within our LISA framework, we conducted a comprehensive ablation study, as summarized in Table~\ref{tab:ablation}. While all ablated variants consistently outperform the ResNet-18 baseline on IV/LBW ($8.02^\circ$/$6.41^\circ$), the full model achieves optimal performance ($\mathbf{7.05^\circ}$/$\mathbf{5.39^\circ}$), indicating that these modules work synergistically rather than independently.

Specifically, removing the \textit{Spectral Injection Block} results in the most significant degradation, particularly in cross-subject generalization (LBW error rises sharply to $6.33^\circ$). This confirms that frequency-domain interactions are indispensable for learning domain-invariant representations, effectively filtering out subject-specific high-frequency noise while retaining stable structural semantics. Similarly, excluding \textit{Spatial Saliency Gating} leads to a notable increase in error (up to $7.25^\circ$ on IV), validating its critical role in suppressing background clutter and focusing the model on informative ocular regions. Finally, the exclusion of the \textit{Semantic Disentanglement Module (SDM)} also incurs a performance penalty. This result validates the necessity of our CLIP-guided orthogonal regularization, thereby preventing the model from learning spurious correlations.

\section{Conclusion}
In this paper, we introduce LISA, a framework designed to address persistent bottlenecks of spatial misalignment and semantic ambiguity in driver gaze estimation. By synergizing frequency-domain consistency via the FAM Fusion module with CLIP-guided semantic disentanglement, our architecture effectively stabilizes features against environmental perturbations and explicitly decouples authentic gaze cues from appearance interference, such as accessories and glare. Extensive experiments on the IV and LBW benchmarks validate that LISA achieves state-of-the-art accuracy while maintaining computational efficiency, demonstrating superior resilience to severe occlusions and lighting changes compared to existing baselines. Future work will explore the extension of this paradigm to the temporal domain and event-based camera modalities to further improve robustness under extreme dynamic conditions.

\section*{Acknowledgements}
This work was supported in part by the Beijing Natural Science Foundation(Grant No.L233034, No.L253004), in part by the National Natural Science Foundation of China (No.62572075) and in part by Fundamental Research Funds for the Beijing University of Posts and Telecommunications (No.2025TSQY01).

\bibliographystyle{named}
\bibliography{ijcai26}

\end{document}